School of Electronics and Computer Science

Faculty of Physical Sciences and Engineering

University of Southampton

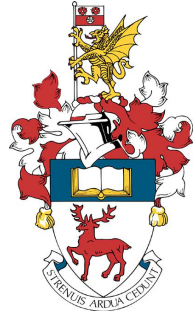

Jakub J. Dylag

May 2nd 2022

# Machine Learning based prediction of Glucose Levels in Type 1 Diabetes Patients with the use of Continuous Glucose Monitoring Data.

Project supervisor: Dr Nicholas Gibbins

Second examiner: Prof Koushik Maharatna

A project progress report submitted for the award of

MEng Computer Science with Artificial Intelligence

## Abstract

A task of vital clinical importance, within Diabetes management, is the prevention of hypo/hyperglycemic events. Increasingly adopted Continuous Glucose Monitoring (CGM) devices offer detailed, non-intrusive and real time insights into a patient's blood glucose concentrations. Leveraging advanced Machine Learning (ML) Models as methods of prediction of future glucose levels, gives rise to substantial quality of life improvements, as well as providing a vital tool for monitoring diabetes.

A regression based prediction approach is implemented recursively, with a series of Machine Learning Models: Linear Regression, Hidden Markov Model, Long-Short Term Memory Network. By exploiting a patient's past 11 hours of blood glucose (BG) concentration measurements, a prediction of the 60 minutes is made. Results will be assessed using performance metrics including: Root Mean Squared Error (RMSE), normalised energy of the second-order differences (ESOD) and $F_1$ score.

Research of past and current approaches, as well as available dataset, led to the establishment of an optimal training methodology for the CITY dataset, which may be leveraged by future model development. Performance was aligned with similar state-of-art ML models, with LSTM having RMSE of 28.55, however no significant advantage was observed over classical Auto-regressive AR models. Compelling insights into LSTM prediction behaviour could increase public and legislative trust and understanding, progressing the certification of ML models in Artificial Pancreas Systems (APS).

## Statement of Originality

- I have read and understood the ECS Academic Integrity information and the University's Academic Integrity Guidance for Students.
- I am aware that failure to act in accordance with the Regulations Governing Academic Integrity may lead to the imposition of penalties which, for the most serious cases, may include termination of programme.
- I consent to the University copying and distributing any or all of my work in any form and using third parties (who may be based outside the EU/EEA) to verify whether my work contains plagiarised material, and for quality assurance purposes.

We expect you to acknowledge all sources of information (e.g. ideas, algorithms, data) using citations. You must also put quotation marks around any sections of text that you have copied without paraphrasing. If any figures or tables have been taken or modified from another source, you must explain this in the caption and cite the original source.

**I have acknowledged all sources, and identified any content taken from elsewhere.**

If you have used any code (e.g. open-source code), reference designs, or similar resources that have been produced by anyone else, you must list them in the box below. In the report, you must explain what was used and how it relates to the work you have done.

**I have used Python, Jupyter Notebooks, PyTorch, hmmlearn, Scikit-learn, Numpy, Pandas, Matplotlib, Seaborn, Pickle. CITY dataset, published by Jaeb Center for Health Research.**

You can consult with module teaching staff/demonstrators, but you should not show anyone else your work (this includes uploading your work to publicly-accessible repositories e.g. Github, unless expressly permitted by the module leader), or help them to do theirs. For individual assignments, we expect you to work on your own. For group assignments, we expect that you work only with your allocated group. You must get permission in writing from the module teaching staff before you seek outside assistance, e.g. a proofreading service, and declare it here.

**I did all the work myself, or with my allocated group, and have not helped anyone else.**

We expect that you have not fabricated, modified or distorted any data, evidence, references, experimental results, or other material used or presented in the report. You must clearly describe your experiments and how the results were obtained, and include all data, source code and/or designs (either in the report, or submitted as a separate file) so that your results could be reproduced.

**The material in the report is genuine, and I have included all my data/code/designs.**

We expect that you have not previously submitted any part of this work for another assessment. You must get permission in writing from the module teaching staff before re-using any of your previously submitted work for this assessment.

**I have not submitted any part of this work for another assessment.**

If your work involved research/studies (including surveys) on human participants, their cells or data, or on animals, you must have been granted ethical approval before the work was carried out, and any experiments must have followed these requirements. You must give details of this in the report, and list the ethical approval reference number(s) in the box below.

**My work did not involve human participants, their cells or data, or animals.**

# Acknowledgements

I would like to thank Dr Nicholas Gibins for supervising this project and mentoring me throughout. His continuous guidance, support and feedback were invaluable and allowed me to achieve this project. Many thanks to Prof Koushik Maharatna for examining my project and providing insightful suggestions.

Utmost thanks to Prof. Michael Boniface, Dr. Chris Duckworth and the rest of the team at the IT Innovation Centre, for the experience gained during my Summer Internship, which I found essential during this project.

Last but not the least, my parents for their patience and encouragement.

# Contents

# Nomenclature

AI - Artificial Intelligence
APS - Artificial Pancreas Systems
AR - Autoregressive
BG - Blood glucose
BMI - Body Mass Index
CF - Correction factor
CGM - Continuous Glucose Monitoring
CITY - CGM Intervention in Teens and Young Adults with Type 1 Diabetes
CR - Carbohydrate-to-insulin ratio
EM - Expectation-Maximisation Algorithm
ESOD - Energy of Second Order Difference
GMM - Gaussian Mixture Modelling
GRU - Gated Recurrent Unit
HMM - Hidden Markov Models
HbA1c - Haemoglobin A1c test
LIME - Local Interpretable Model-agnostic Explanations
LRP - Layer-wise relevance propagation
LSTM - Long Short-Term Memory
ML - Machine Learning
MSE - Mean Square Error
PH - Prediction horizon
PMCC - Product Moment Correlation Coefficient
RMSE - Root Mean Square Error
RNN - Recurrent Neural Networks
S.D. - Standard Deviation
SHAP - Shapley Additive Explanations
T1D - Type 1 Diabetes

# 1 Introduction

Artificial Intelligence has made large advances in preventive medicine and management of chronic conditions, such as Diabetes. The symptoms of this chronic, metabolic disease include unregulated blood glucose (BG) levels and hyperglycaemic events, which over time can lead to irreversible damage to the heart, kidneys, blood vessels, eyes and nerves.

Type 1 Diabetes (T1D) Patients must use external insulin, with effective management requiring optimal doses of insulin, through injection or infusion, multiple times per day. Current technologies have found success in minimising risk of long-term complications, however can also pose a significant burden on self-managing patients, reducing quality of life.

Recent growth in digital disease management platforms and an increased use of diabetes-related data sensors, such as Continuous Glucose Monitoring (CGM), have resulted in rapid evolution towards Autonomous Diabetes Systems, which implement Machine Learning (ML) approaches to predict future glucose levels, detect hyperglycaemic events, assist with decision making and risk stratification for T1D patients.

The implementation of leading Neural Network based ML models could advance prediction accuracy and provide more personalised predictions. This leads to the opportunity to explain model predictions and justify conclusions reached, crucial for medical use.

# 2 Literature Review

## 2.1 Diabetes

Diabetes Mellitus affects the pancreas' ability to regulate insulin. The condition is categorised into different forms dependant on the cause: Prediabetes, Type 1 Diabetes, Type 2 Diabetes, Gestational Diabetes [4]. The International Diabetes Foundation estimates there are over 463 million diabetics worldwide, with projected increases to 578 million by 2030 [3]. The Insulin hormone is responsible for intake of glucose into muscles, effectively decreasing BG levels. When digesting carbohydrates, glucose is introduced to the bloodstream, risking excessively high levels (hyperglycemia). A bolus of insulin may need to be taken to reduce glucose levels, or before performing exercise.

T1D is an autoimmune condition thought to be caused by the destruction of insulin-producing beta type cells [2] by the immune system. Symptoms of marked hyperglycemia include polyuria (excessive urination), polydipsia (extreme thirst), weight loss, sometimes with polyphagia (extreme hunger), ketonemia and blurred vision [5]. Hood and Colleagues [6] also indicated increased risk of elevated depression symptoms in children and adolescents. Although the condition can appear at any age, it is most common to appear around the ages of 4-7 and 10-14 [1]. Historically, T1D diagnosis has included fasting BG levels higher than 126 mg/dL or any measurement of 200 mg/dL. Diagnosis can also be made with a HbA1c (average BG concentration over past 3 months) value over 864 mg/dL mmol/mol [2]. Despite more recent efforts to standardise diagnosis, causes and typology remain unclear among adults [2].

Although patients who self-monitor BG levels can see long-term benefits, limitations such as inconvenience, poor adherence to frequent measurements and financial costs of disposables may result in erroneous readings [7] and therefore inaccurate predictions. Recent advances in sensor technology enabled CGM devices to be developed. CGM devices greatly reduce feelings of burden and of inconvenience [10]. These sensors are minimally invasive, using a single electrode needle and perform readings every 1-5 minutes for consecutive days or weeks [9].
Juvenile Diabetes Research Foundation [8] demonstrated reductions in nocturnal hypoglycaemia in children and adolescents when using CGM compared to self-monitoring. The increasing amount of frequent data is causing the decision-making process to become more complex. Current research is focused on developing algorithms to simplify this process and progressively automate in the future. Prototypes of closed loop Artificial Pancreas Systems (APS) [11] are being investigated by the open source community, which integrate CGM devices with Automatic Insulin Injection, enabling fully automated therapy with no input from the patient or medical staff.

The range of Autonomous Disease Management systems has been classified by Bitterman and colleagues [22] into a set of 5 levels of autonomy, each with specific risk management and legal liability, as presented in Figure 1 below.

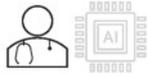

Figure 1: levels of autonomy [22]

CGM devices alone are limited to data presentation functionality, however with addition of predictive algorithms are classified as level 3. Future APS will have higher or full degrees of automation, introducing legal responsibility for the AI developers.

## 2.2 Past Approaches

Early Decision Support Systems are designed to help medical professionals in retrospective analysis of patient's data. Centralised Telemedicine services were enabled by remote communication, through integrations of Web and Telephone access, between Patient Units and central Patient Management Units [13, 14]. This allowed for the implementation of empirical mathematical models that calculate the new dose of the medication required to achieve therapeutic goals [15].

A standard T1D therapy includes the use of a Bolus Calculator to calculate the dose of insulin bolus required using the equation:

$$B = \frac{CHO}{CR} + \frac{G_C - G_T}{CF} - PS \times IOB$$

*Equation 1: Bolus Calculator [12]*

The first part of the formula is concerned with finding the meal boluses required to offset ingested carbohydrates during the meal. This is calculated as the ratio between carbohydrate intake (CHO) and carbohydrate-to-insulin ratio (CR), a parameter specific to each patient.

Next the correction insulin dose to alter BG concentrations to the target level is calculated dividing the difference between measured concentration ($G_C$) and the target concentration ($G_T$), by the correction factor (CF).

Finally, to avoid insulin stacking from the previous injection subtraction of the Insulin on Board (IOB) is necessary. The multiplier physiological state (PS) can be introduced to increase accuracy. PS is <1 when insulin sensitivity is increased, for example during exercise and PS >1 when sensitivity is decreased, for example during illness [16].

Despite standardisation and the success of these empirical approaches, prediction accuracy is limited by the high degree of generalisation.

## 2.3 Current Approaches

Within this section more complex and personalised methods of glucose management will be investigated.

Adaptive Tuning of Bolus Calculator Parameters could better personalise therapy and automate periodic adjustments by clinicians, reducing the cost of care.
Herrero and colleagues [17] automatically adjusted the CR and CF parameters everyday in a run-to-run methodology based on CGM data. Although results on simulated data were promising, with significant reductions in time spent in hypoglycaemia, a fundamental flaw exists in the assumption of repetitive BG levels. Many factors such as exercise, hormone cycle, alcohol, stress, mental illness and others, can dramatically affect glucose levels and are not accounted for.

With the addition of case-based reasoning to the run-to-run methodology situational context can be comprehended by the algorithm. Successful recent approaches from Sun and colleagues [18] incorporate reinforcement learning and produced encouraging results on virtualized data by personalising parameters for better glycemic control even in extreme scenarios. However this has not been thoroughly tested with real CGM data.

CGM devices not only allow for detection of hypo/hyperglycemic events, but also permits prediction. There are two main approaches in the literature. Firstly, viewing it as a regression problem and predicting the values of BG concentration into the future until a prediction horizon (PH). Secondly, forecasting the occurrence of future hypo/hyperglycemic events [9].

Another distinction can be made on the origin of training data. Some algorithms only use CGM data while others implement external input for meal content and amount of injected insulin, which have been shown to enhance predictions [20]. Furthermore, given the limited amount of CGM data, some studies use simulated data (in silico trial) from the widely used UVA/PADOVA simulator [21].

Population algorithms are standardised and reused throughout the whole population. Although this approach can leverage large CGM datasets and reduce development time, by only needing to train a single model, predictions are not personalised to the patient.
By contrast, subject-specific algorithms which orchestrate various specialised models to produce a combined prediction, account for the variability between individual's characteristics [19].

Prediction of glucose concentration models have a limited PH under 60 minutes. This allows for the patient to be alerted up to 60 minutes before the glucose concentrations become too high/low. As presented in Figure 2 below, predictions enable patients to pre-emptively digest carbohydrates or inject insulin prior to surpassing dangerous thresholds, minimising the time spent hyper/hypoglycemic.

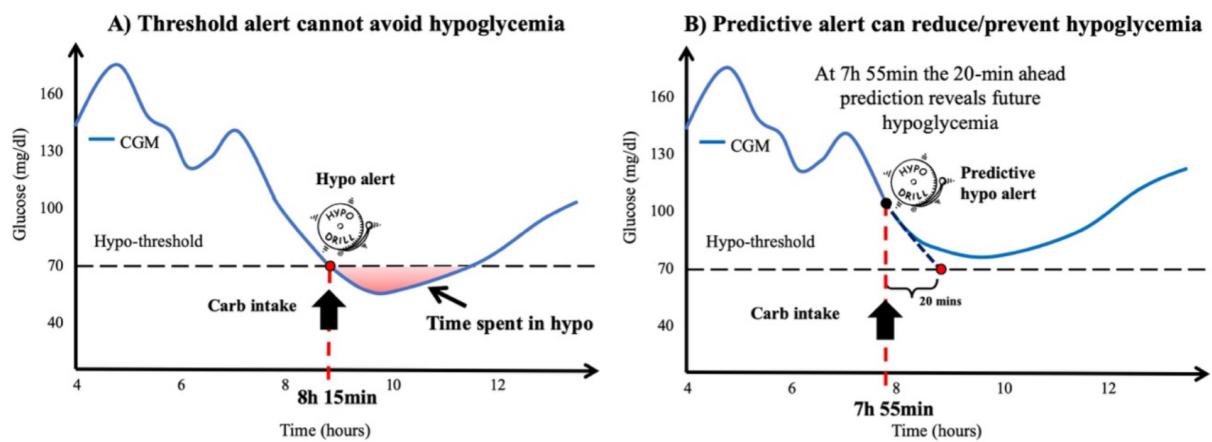

*Figure 2: predictions made on nocturnal CGM series plot [9]*

Event prediction models reduce prediction difficulty by only forecasting the time of occurrence without the exact glucose concentration levels. This simplification enables a larger PH between 2-3 hours, giving the ability for the patient to identify postprandial hypoglycaemia at the time of eating, allowing the patient to not have to eat extra carbohydrates, managing their weight and promoting a healthier lifestyle.

## 2.4 Machine Learning

The most recent CGM sensors [23] adopt simple prediction algorithms such as linear extrapolation 15–30 min in advance and generating alerts if a hypo/hyperglycemic threshold is predicted to be surpassed [24]. Accuracy can be further improved by utilising LASSO (L1) regularisation [29].

Autoregressive (AR) models capture the signal's frequency information and are invariant to the signal's phase and amplitude [25]. AR is a popular approach for both Adaptive Tuning of Bolus Calculator Parameters, as well as regression like glucose level prediction. The simplified parameters allow for detailed interpretations, in contrast to more complex models described below.

### 2.4.1 Hidden Markov Models

Hidden Markov Models (HMM) are an extension of the Markov process [26]
This state-space model in which latent variables represent discrete values, is considered to be a double Stochastic process, consisting of Hidden States that cannot be observed, in addition to a random sequence of observations dependent on previous states [39]. This property makes them capable of predicting and analysing any temporal sequence.

Defined by the number of Hidden states (N), as well as number of observation symbols per state (M). Model parameters dictate the state transitions and are estimated by a special case of Expectation-Maximisation (EM) Algorithm, known as the Baum-Welch algorithm [40]. Of note is that this algorithm is a gradient-based optimization method, which is vulnerable to stagnation in local maximas [41]. Other recent approaches implement both supervised and unsupervised training of HMM.

The Veterbi algorithm is most commonly used to infer optimal hidden states, hence predicting the next time step. A decoder is also necessary to map the state-space of discrete random variables to an comprehensible output value. By increasing the number of hidden states output resolution is enhanced however at the expense of model complexity; leading to increased computation and data requirements.

HMM has been demonstrated to be optimal for small training sets (N < 1000) [26].

### 2.4.2 Long-Short Term Memory

Introduced by Hochreiter and Schmidhuber in 1997 [45] Long-Short Term Memory (LSTM) networks are a leading accuracy in complex time series prediction tasks [28], by leveraging its capability to bridge long-term dependencies in excess of 1000 time steps [45].

LSTMs are well suited for temporal data and have shown advantages in Machine Translation [56], Speech recognition [57] and mortality prediction in Intensive Care Units, based on Electronic Healthcare Records [58].

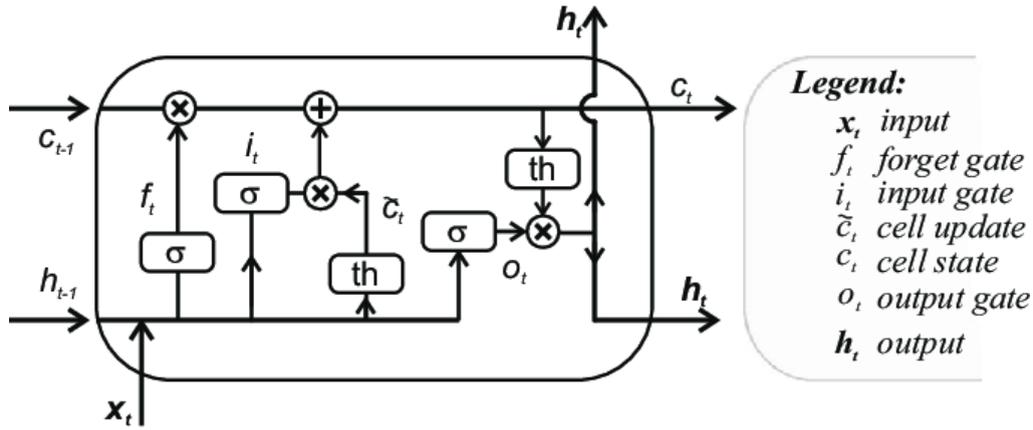
*Figure 3: Graphic illustration of LSTM cell [51]*

LSTM cells hold 2 types of memory: short and long. Hidden States ($h_t$) are used throughout Recurrent Neural Networks (RNN) to store immediate previous events and are constantly majorly altered by a series of gates. The internal Cell State stores long term dependencies, with only minor linear interactions throughout the forward pass [44]. This enhances RNN performance, which is inhibited by the vanishing gradient problem [53]. Both are represented as vectors of predefined size, with larger vectors increasing model complexity.

Both the cell and hidden states are fed back to the input of the cell, alongside the next time step and the calculations are repeated until the whole sequence is processed. Once the final time step is reached the terminal cell state is passed to a fully connected output layer, which subsequently produces the final output [43].

$$\begin{aligned}
i_t &= \sigma(W_{ii}x_t + b_{ii} + W_{hi}h_{t-1} + b_{hi}) \\
f_t &= \sigma(W_{if}x_t + b_{if} + W_{hf}h_{t-1} + b_{hf}) \\
g_t &= \tanh(W_{ig}x_t + b_{ig} + W_{hg}h_{t-1} + b_{hg}) \\
o_t &= \sigma(W_{io}x_t + b_{io} + W_{ho}h_{t-1} + b_{ho}) \\
c_t &= f_t \odot c_{t-1} + i_t \odot g_t \\
h_t &= o_t \odot \tanh(c_t)
\end{aligned}$$

*Equation 2: Formulaic definition of LSTM cell [46]*

Of note is the forget gate ($f_t$), which is a sigmoid layer selecting what information to discard from the cell state. The input ($i_t$) and output ($o_t$) gates transform the previous hidden state ($h_{t-1}$), using a sigmoid activation function, to incorporate into cell state ($C_t$) and output hidden state ($h_t$) respectively.

LSTM flexible architecture offers a range of possible configurations.
Bidirectional LSTM (BiLSTM) architecture is similar to vanilla LSTM, however the sequence is processed in both the forward and backwards direction, which requires the entire sequence to be available at once.

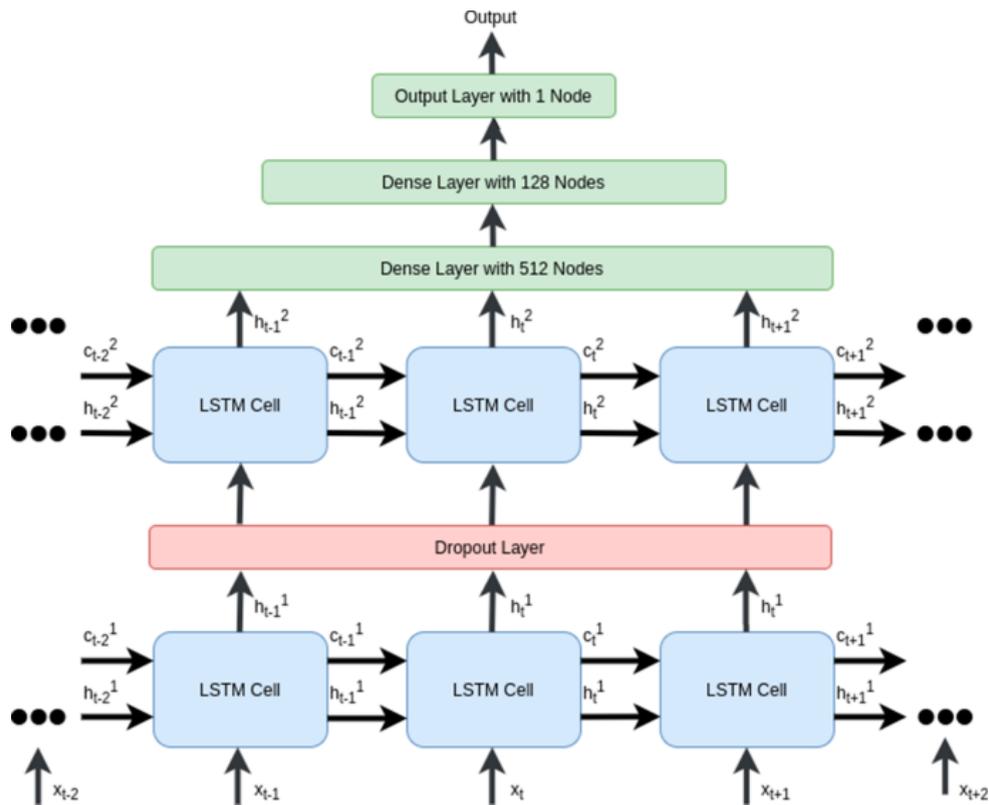
*Figure 4: Illustration of Stacked LSTM architecture* [52]

The Recurrent nature of LSTM enables the forwarding of the cell state, not only to the next cell, but also to a deep LSTM layer. These layers take the cell state from previous layers as input as opposed to the input data, which allows for increasing levels of abstractions to form.

With these increased capabilities, multiple parallel series can be effectively analysed, forming a single prediction. This can be utilised as a method of inputting static variables by creating a constant time series for each. However complexity will be increased, strengthening requirements for more layers and data to achieve sufficient accuracy.

Convolutional LSTMs (ConvLSTM) incorporate a one dimensional convolutional layer previous to the initial LSTM layer [54], in order to allow for deeper understanding. Renzhuo Wan et al. [55] has demonstrated ConvLSTM superiority in multivariate time series prediction over other attention models. Although BiLSTM has lowest RMSE in Sequence-to-Sequence prediction, ConvLSTM marginally improves upon it for individual models [42].

Other variants of LSTM include Gated Recurrent Unit (GRU) [59] and the addition of peephole connections [60]. GRUs simplify the LSTM architecture and reduce training cost by combining the input and output gates to form a single update gate. However LSTMs remain prevalent in most use cases.

A challenging problem with LSTM architecture most prevalent with stacked LSTMs, is overfitting. A solution to overcome this is the use of Drop Out layers in between LSTM layers for better generalisation, as seen in figure 4. This randomly discards node outputs, altering the following layers perspective and entices generalisation.

In contrast to HMM, LSTM harnesses greater accuracy using large datasets (N > 1000) [26]. However LSTM complexity is detrimental to prediction accuracy when using smaller datasets.

## 2.5 Model Explainability

Clinical practitioners and patients would like more granularity in alerts and to better understand the reasoning behind it. Excessive risk notifications could lead to alert fatigue, where patients or doctors become desensitised and are less likely to take necessary action.

Current algorithms, such as Shapley Additive Explanations (SHAP) and Local Interpretable Model-agnostic Explanations (LIME), focus on extracting feature importance. SHAP conditions on each feature, measuring change in prediction based on given changes in input feature [63]. LIME samples neighbouring nodes in the region of interest and approximates their behaviour with simpler surrogate functions [62]. Although demonstrating the importance of a feature is vital in large multi-feature Neural Networks, temporal datasets with limited features do not benefit from this approach. Instead it is crucial to survey event importance in order to extract meaning from time-series prediction algorithms.

A common approach is Layer-wise relevance propagation (LRP) [30], which interprets individual predictions in terms of input variables, by propagating the predictions backwards. This method. It has been used to provide for state-of-the-art models such as VGG-16 [30] and can be effectively applied to any Neural Neural or LSTM model. [31]
When applied to LSTM networks, LRP procedures can be restricted to only utilise the necessary many-to-one weighted linear connections and two-to-one multiplicative interactions evident in LSTMs [61]. One primary problem with LRP is that it requires the definition of target classes, which is not suitable for continuous predictive models.

A specific method of analysing LSTM would be logging the forget gate, which controls the importance of the previous cell state (as demonstrated in Section 2.4.2). Research has shown that the forget gate is one of the most important gates in the LSTM, with forget-gate-only versions outperforming standard LSTM models on certain benchmark datasets [64]. Higher values of the forget gate vector indicate that the previous cell state needs to be remembered, as opposed to lower values that justify forgetting the previous cell [65]. By recording the state of the forget gate during each prediction at the inference stage, the importance of the previous cell state can be viewed as the importance of a single time step compared to the sequence as a whole.

# 3 Methodology

## 3.1 Data Acquisition

Identifying a suitable CGM T1D Dataset is vital to the success of this project.
In addition to maximising the number of measurements, consistency within the dataset is essential. Minimal interruptions and gaps, minimise interpolation required, enhancing the quality of the data and true representation of trends. To add to that, a varied study cohort is required to ensure good representation of population and promote generalisation in the model.

Various Datasets [32, 33, 66, 67] have been found, however due to the privacy restrictions in the UK/EU the search was limited to public datasets in the US.

| Dataset | Patients | Age | Duration |
| --- | --- | --- | --- |
| Tamborlane 2008 | 451 | 8 > | 6 months |
| Chase 2005 | 200 | 7-18 | 6 months |
| CITY 2019 | 153 | 14-25 | 22 months |
| Aleppo 2017 | 224 | 25-40 | 6 months |
| OhioT1DM 2020 | 12 | 20-60 | 2 months |
| Maastricht 2010 | 197 (T2D) | 40-57 | 3 months |
| Weinstock 2016 | 200 | 60 > | 14 days |

*Figure 5: Tabular presentation of most prominent surveyed Datasets*

OhioT1DM Dataset [66] commonly used in other studies [36], includes 8 weeks of data collected from 12 anonymous T1D patients, between the ages 20 and 60 years old. Data was highly unsynchronized as collected from multiple devices and sometimes logged manually [66].

Most datasets focus on a specific age group, enabling a combination of 5 datasets to form a complete lifespan sequence. Although joining datasets could expand the size and scope of the model, the accuracy and usage between different models of prevalent CGM devices, as well as environmental factors and testing methodology differences, can have a detrimental impact on prediction accuracy, given increased variability of data quality.

CGM Intervention in Teens and Young Adults with T1D (CITY) Dataset [32] was chosen, featuring data from 153 patients over a period of 22 months [34]. With the largest amount of CGM data collected from similar CGM devices (Dexcom G4/G5/G6). Using a large dataset would allow for more flexibility in model choice, training methodology and more accurate testing.

*The source of the CITY dataset is Jaeb Center for Health Research, but the analysis, content and conclusions presented herein are solely the responsibility of the authors and have not been reviewed or approved by Jaeb Center for Health Research

## 3.2 Data Analysis

### 3.2.1 Patient Cohort

CITY focuses on teenagers and young adults, with a mean (s.d.) age of 17.46 (2.91), with 100 patients between the ages of 14 and 18 (inclusive) and 53 over 18, at the time of enrollment. The maximum age was 24.

Insufficient frequency of patients' physical examinations, lead to the decision of representing variable physical features as static by computing the mean or max as appropriate.
The mean (s.d.) weight was 74.26 kg (16.22) and 169 cm (10.28) for height. Therefore the mean (s.d.) Body Mass Index (BMI) was 25.96 (5.14), which means 50.9% of patients are classified as atleast overweight with 17.6% obese. In comparison with the United States adult average BMI of 29.3, with 71.0% being at least overweight [68], the studies cohort is shown to be healthier on average.

Commonly expected biological trends, such as on average men being taller and heavier than women, were represented in the population. Interestingly, the frequent assumption of a positive correlation between weight and resting heart rate was absent. The same is true for abnormal health judged by a detailed medical assessment, which was randomly distributed within the population, independent of both weight and height. It is important to note that correlations observed from this limited sample are not representative of the population at large and should be taken with great caution.

Further Analysis of the patient population indicated patients were highly educated with over 74% attending university. Ethnicity was highly homogeneous with 129 out of 153 (84.3%) patients being of white background. Annual Income in US Dollars had a mean (s.d.) of $64,869 (58,348). This elevated standard deviation, alongside a high variance of 3,404.51, quantifies a large division in annual income, with 36 patients earning over $100,000 and 41 earning less than $25,000. Results are consistent with the median household income of $67,521 in 2020 [69].

Despite the scope of the data limited to teenagers and young adults with T1D, selection bias in the study cohort is evident towards the white ethnic group and the highly educated. This cohort may not be representative of common glucose concentration behaviour, therefore limiting the generalisation and adaptability of the model.

### 3.2.2 CGM Measurements

Of note are the Florida USA origins of the data, where the unit for BG concentration is mg/dL, whereas the European standard is mmol/L (1 mmol/L = 18 mg/dL). CGM Readings were distributed in a positively skewed normal distribution, with a mean of 204.56 mg/dL and standard deviation of 87.21. Interestingly, the largest concentration of reading is at 401 with 3.59%. This could be due to the limited range of the CGM devices (Dexcom G4/G5/G6). However further extreme readings exist, with 1009 readings (0.012%) above 401 mg/dL up to a maximum of 600 mg/dL, which is diagnosed as a Diabetic Coma [70]

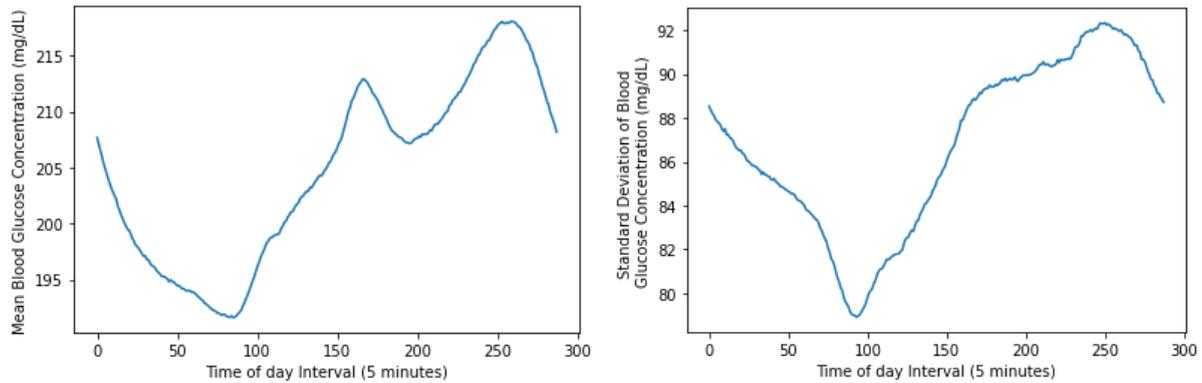

*Figure 6: Mean (left) and s.d. (right) of daily CGM readings grouped at 5 minute intervals*

Minimum average levels of 191.61 mg/dL can be observed at 7:05 with maximums of 218.08 at 21:40. Variability of readings remains very high at a minimum of 78.92 across the whole day, with further increases up to a maximum of 91.08 during evening hours. This demonstrates instability of BG levels and large variability between consecutive days (weak cyclical behaviour), which can be medically correlated to the patient's exercise regime and diet throughout the day. Although this posed greater difficulty of prediction, it allowed for more flexibility during training set creation.

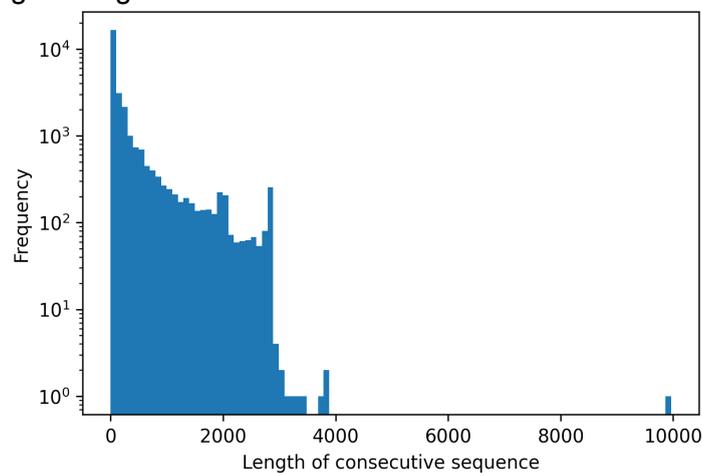

*Figure 7: Distribution of Lengths of consecutive CGM reading sequences*

Readings were not performed consistently, with unpredictable periods of missing data as a result of multiple factors such as: CGM device battery depletion, patient forgetting to apply CGM device. This is quantified by a large standard deviation of 562.33. Half of the sequences (N=14,079) were under 50 in length, with all sequences were under 4,000 (13.89 days) in length, except a single extrema of length 9,963 (34.59 days).

### 3.2.3 Statistical Methods

Statistical Analysis of Patient Data was conducted in order to select essential static features, in addition to temporal CGM readings.

Covariance Matrix and Correlation Matrix were computed with all values normalised previous to calculations to eliminate value range bias. The covariance quantifies the direction of the relationship, with a positive covariance signifying a positive correlation and similarly a negative value for a negative correlation. Product Moment Correlation Coefficient (PMCC) neatly accompanies the covariance matrix and is used to calculate the strength of a correlation. Leveraging both matrices allows for the identification of both the direction and strength of trends between features.

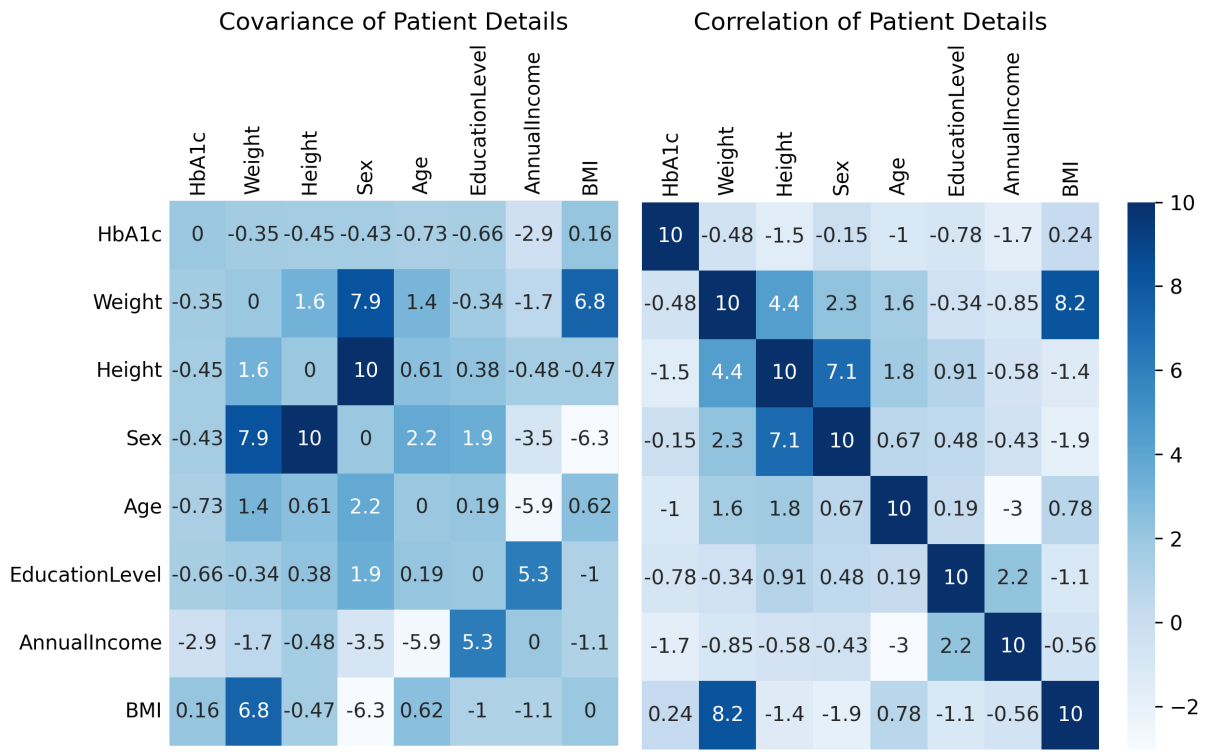

*Figure 8: Covariance and Correlation Matrix for Patient Details (normalised between 10 and -10 for illustration purposes)*

HbA1c has a high variance ($10.97 \times 10^{-3}$) indicating the high variability in diabetic patients, which could be detrimental to generalised prediction models.

Furthermore HbA1c covariance with weight and height are significantly reduced, with $-6.33 \times 10^{-4}$ and $-8.14 \times 10^{-4}$ respectively, in addition to further reductions with BMI. Interestingly HbA1c and BMI are positively correlated (0.024), in contrast to the negative correlations with Weight and Height (-0.048 and -0.151). This minor correlation goes against common belief, presenting negligible correlations between BG levels and primitive health indicators such as BMI, suggesting independence between body fat percentage and BG concentrations.

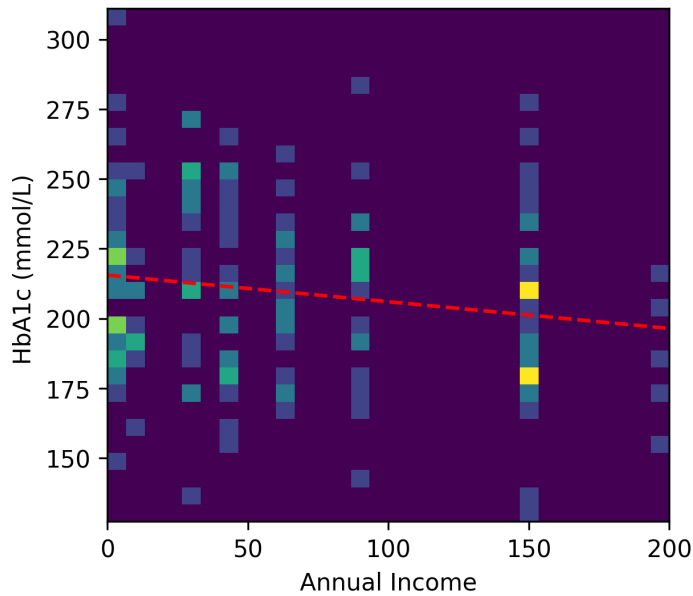

*Figure 9: Illustration of Annual Income ($10^3$ USD) against HbA1c (mg/dL) (m = -0.0955)*

Analysing the correlation between HbA1c and Annual income presents a significant causal relationship. Elevated PMCC between the two (-0.17), in addition to negative linear correlation (presented above) reinforces the theory of lifestyle and dietary differences between social classes impacting patient glucose levels. This could be attributed to the increased cost of low glycemic food sources such as Protein and Vegetables, in comparison to cheap processed foods, prohibiting access to lower income patients and elevating their exposure to sugar-rich foods.

Most important features can be selected using variance thresholding [71] approach, which removes all variables whose variance does not exceed threshold. This yields the most meaningful features to be: Annual income, Education Level, HbA1c, Height and weight. However as BMI is calculated from the combination of height and weight, it is a more meaningful feature and therefore better representative of patient health.
HbA1c, Annual income and BMI are concluded to be the most crucial features.

## 3.3 Clustering

Calculations of cohort variability revealed major divergences within the patient population, advocating leveraging these differences by clustering patients into distinct cohorts, thus decreasing variability in patient characteristics. Statistical methods quantified the large covariances within patient details (Section 3.2.3). Correspondingly, CGM data was found to have weak cyclical behaviour and extremas (Section 3.2.2). The aforementioned indicates the scope of major divergences in the population.

An assumption can be made that contrasting patient characteristics lead to heterogeneous behaviour of BG concentrations. In the case of similar patient characteristics a generalised model trained upon a single population could be successfully implemented. By contrast, with large variations between patients it may be necessary to train personalised models for each patient, which could restrict the scope of CGM behaviour, decreasing prediction difficulty. However this would significantly reduce the size of the training set, introducing a prevalent risk of underfitting in complex models, such as LSTM.

Gaussian Mixture Modelling (GMM) [72] was chosen over K-means clustering, which assumes circular distribution by using the euclidean distance between points, whereas GMM allows for the elliptical spread of data for better division. The expectation-maximisation (EM) algorithm [73], was utilised to optimise centroid positioning. Insurances were made that the solution was optimal by performing 200 iterations for each of the 20 random initial settings.

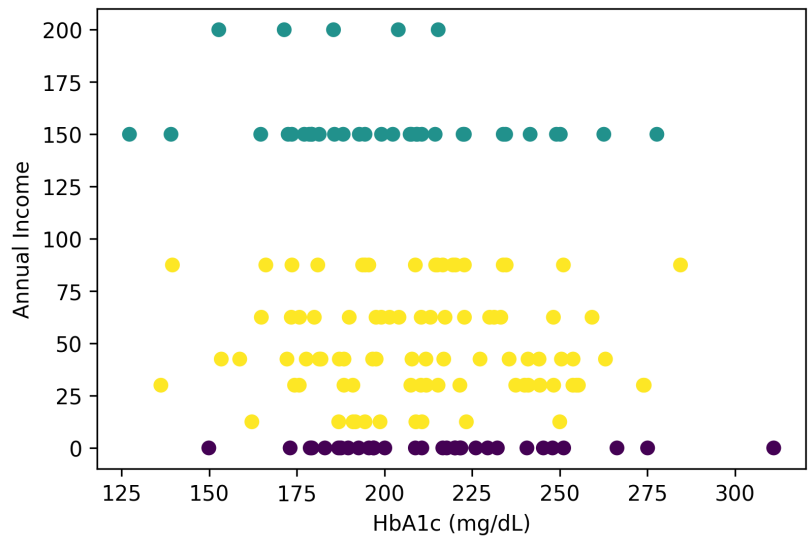

*Figure 10: Relationship between HbA1c (horizontal) and Annual Income(vertical) across the coloured 3 cohort of patients.*

Effectiveness of groupings could be quantified by the lower bound value on the log-likelihood of the optimal clustering. Although combinations of most varianced variables such as sex and annual income would provide the most distinct groupings, it is evident that these variables would have the least impact on oscillations of CGM readings, which would not substantially improve model performance. Experimentation with a range of more meaningful variables directed to the choice of HbA1c and Annual Income with 3 distinct cohorts, resulting in the lowest log-likelihood of -8.25.

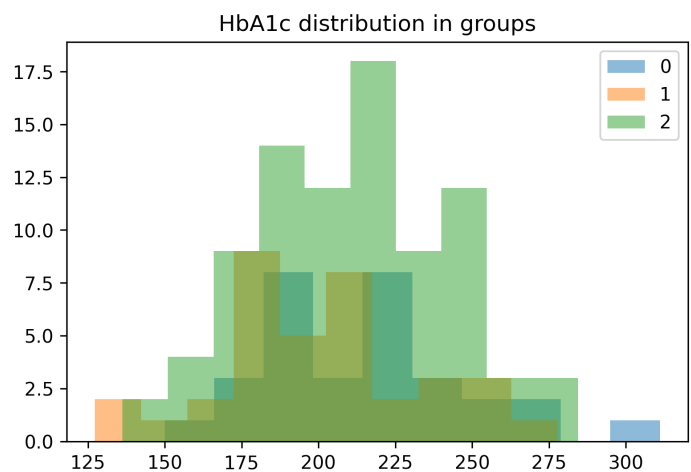

*Figure 11: Distribution of HbA1c across 3 cohorts.*

Further analysis was conducted into the resulting distributions of variables for each cohort, visually illustrating the distinctness of the clustering. It is evident that GMM was focused on the separation of annual income brackets with high, medium and low groupings. However HbA1c has significant overlap with limited distinction, which could be detrimental to model performance.

## 3.4 Design Methodology

Although Hypo/Hyperglycemic Event Forecast Models could give greater benefit to the patient with longer PH, Glucose Concentration Prediction Models are more widely used, offer a deeper insight into the explainability of the model's prediction process, as well as potential use in Artificial Pancreases in the future.

Such regressive models for Glucose concentration prediction can be used both in a recursive and direct fashion. Recursive methods incorporate the previous prediction in their input, whereas direct methods do not [36]. Direct methods require the implementation of sequence to sequence models (many-to-many), which inflate model complexity exacerbating underfit risk. Recursive predictions lead to the accumulation of error, resulting in a lesser accuracy, however deeper explainability insights can be gained, by analysis of each individual time step. More thorough analysis could promote faster adoption, in addition to trust and reliability for patients and medical staff. Therefore a recursive approach will be implemented.

Analysis of varied Machine Learning Models (Section 2.4) has shown a range of trade-offs between levels of model complexity. Therefore, it is most appropriate that a diverse span of models be used, allowing for detailed comparison and selecting the highest performing. Comparative investigation led to the choice of: Linear Regression, HMM and Stacked LSTM models. Prevalent AR models will not be implemented as previous studies have indicated disjointed sequential data, as quantified in Section 3.2.2, can severely impact accuracy [74]. Although a range of LSTM configurations, such as Bi-directional and Convolutional are available, they significantly impede explainability investigations, as well as increasing the risk of underfitting. Therefore a Stacked LSTM will be implemented enabling flexibility in model complexity, by adjustment of stacked layers.
This chosen range of models enables a thorough comparison, given the same training data and allows for better judgement of accuracy. Discrepancies during comparisons with other predictive solutions for Autonomous Diabetes Systems, could arise from the scarcse use of the CITY dataset, not allowing for equivalent judgement. Thus, internal model evaluation and use of multiple models can help alleviate the aforementioned problems.

A multitude of two training methodologies will be implemented in a separate parallel form. A generalised model will employ all available temporal data, in distinction to a clustering approach, which will develop 3 independent models from each of the cohorts outlined in Section 3.3. Both approaches will be independently trained on a sequence of 132 measurements, which approximates 11 hours, and predict the next 12 measurements (60 minutes). These will be tested against the validation set consisting of 12 readings. Insights from Section 3.2.2 informed this decision, as 10,379 sequences (36.69%) of length more than 144 will be available for training and testing. Although decreasing the required length would allow for more training data, shorter sequences may not supply adequate context to LSTM's long-term memory impacting prediction performance.

It is vital that all CGM sequences belonging to a certain patient must be kept in the same cluster to ensure there is no contamination between the training and testing data sets. Standard randomised sampling of sequences for cross-validation method (Section 5.1), would mix readings between clusters, therefore train-test split must be conducted subsequent to clustering.

Performance will be evaluated using a series of metrics: Root Mean Squared Error (RMSE), $F_1$ score and Energy of Second Order Difference (ESOD). Details are outlined in Section 5.1.

HMM with fewer hyperparameters facilitate more detailed and thorough interpretation, as opposed to LSTM, which are challenging to comprehend. However due to time constraints explainability analysis will be limited to the best performing model. This will aim to quantify the attention placed on each datapoint, emphasising significant events.

# 4 Implementation

## 4.1 Data Preparation

Raw data stored in text-file format, from the CITY dataset, was imported and passed through a preprocessing pipeline for training and testing set creation. Python packages such as Pandas and Numpy were heavily utilised within the pipeline, in addition to all statistical analysis documented in previous sections.

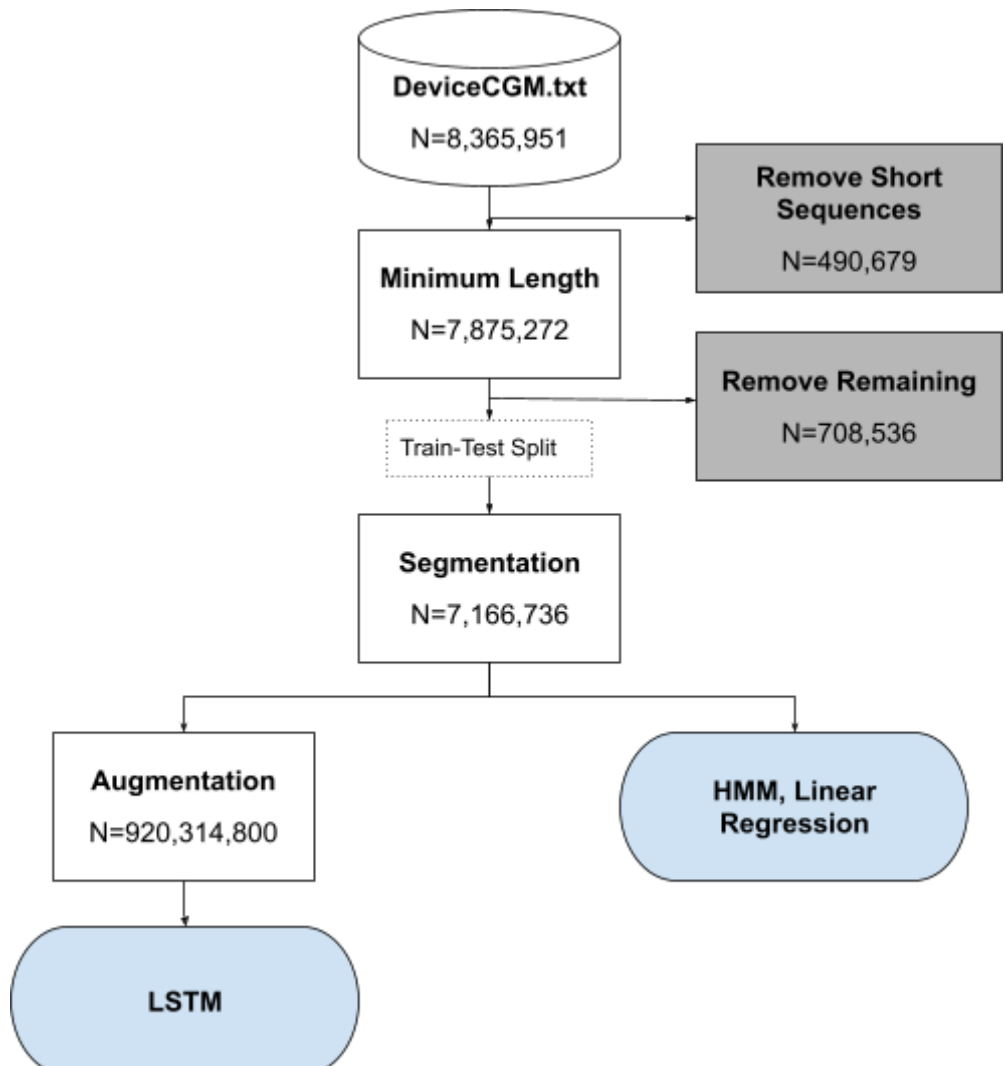

*Figure 12: Illustration of CGM data loss (number of readings) throughout Preprocessing pipeline*

As cohorts were divided by patient, all CGM data from one patient was included in the cohort they belonged to. The same process is computed for the generalised dataset and each of the cohorts, divided from clustering.

Consecutive measurements within sequences must be made within 15 mins, allowing sufficient time for users to switch CGM devices once battery is depleted. 28,290 sequences of connected measurements were identified. However only 10,379 sequences (36.69%) were more than 144 in length (Section 3.4). Subsequently randomised division into test sets and training is performed using a 5-fold cross-validation methodology.

Segmentation into lengths of 144, of both training and test sets was performed (as specified in Section 3.4), with the remaining readings being discarded. The first 132 readings were used as input data with the following 12 used as validation during training. This data was used to train the HMM and Linear Regression models.

Additional data augmentation was performed on LSTM training data, due to the inefficiency of training HMM models. Therefore it was impractical to train HMM with such a large dataset. Augmentation vastly enlarged the number of sequences from 49,769 to 6,391,075, using a sliding window size of 144 samples (11 hours) and step size of 1 (5 minutes). Similarly 132 initial readings are used to train with the following 12 used as validation. Although duplication of sequence segments will not provide unseen information, variations of sequences present varied context, maximising the available training data. Also it is vital to note augmentation was only performed on the training set and not validation set in order to avoid test set contamination and duplication of test sequences producing unequal representation.

| Dataset | Patients | Set Size (sequences) | Training Set Size (sequences) | Test Set Size (sequences) |
|---|---|---|---|---|
| All | 153 | 6,391,075 | 5,112,860 | 1,278,215 |
| Cohort 1 | 86 | 3,296,488 | 2,637,190 | 659,298 |
| Cohort 2 | 36 | 1,203,470 | 962,776 | 240,694 |
| Cohort 3 | 31 | 1,891,117 | 1,512,893 | 378,224 |

*Figure 13: Tabular comparison of all processed datasets used*

Cohort 1 is the largest with 52% of data, followed by Cohort 2 and 3 which roughly share the remaining data at 19% and 30% respectively.

Pandas' Dataframes enabled versatile analysis of given features, however performance limitations exist and Numpy Arrays were favoured for training. Serialisation of the arrays using the python pickle library was conducted in order to limit computation and ensure data consistency between models.

## 4.2 HMM

HMM model was created using the hmmlearn python library [41]. Regular step intervals in the dataset promoted the use of a Discrete HMM as opposed to Continuous HMM, which considers irregular measurements [77]. This would provide further context and disproportionately favour HMM over LSTM. An implementation of Viterbi Training Algorithm [76] was required, as it was not included in the python package. This dynamic programming algorithm approximates the maximum likelihood explanation, predicting the next state, which corresponds to a segment of the possible glucose concentrations. The initial formula $\mu(X_0) = P[Y_0|X_0]P[X_0]$ is utilised to identify the initial best state, given observed data. All further best states (k) are recursively identified using $\mu(X_k) = \mu(X_{k-1})P[X_k|X_{k-1}]P[Y_k|X_k]$. Scaling of the Viterbi output was required to map the ordinal state values to real glucose measurements.

The default Baum-Welch Algorithm [75] implementation in the hmmlearn package was utilised to train the HMM. Training was very slow, with limited parallelisation and no GPU support. All computation was performed on a single CPU core (4.5GHz) with 32 GB of RAM.

| Number of States | Iterations | RMSE |
|---:|---:|---:|
| 3 | 1k | 156.04 (18.07) |
| 20 | 2k | 146.15 (17.92) |
| 100 | 10k | 112.34 (15.73) |

*Figure 14: Table of all HMM models used in hyperparameter tuning*

An array of hyperparameter combinations were investigated. Increases in the number of states led to additional complexity in the model, requiring further iterations to fully train the HMM. The best performant model had the included the most states and iterations, with 100 states and 10,000 iterations resulting in an RMSE of 112.34. Larger HMM models were impractical to train, given that 10,000 iterations took over 12 hours of computation.

## 4.3 LSTM

Creation of the LSTM model was performed in an object-oriented approach permitted by the PyTorch framework in Python. By leveraging polymorphism standardise LSTM PyTorch classes can be inherited and adapted to specific dataset requirements, whilst remaining compatible with other supporting logic through standardised interfaces.

The nn.Module base class [48] for all neural networks, must first be inherited to enable interactions with other modules by implementing the interface through the __init__ () and forward() functions. Specialised layers such as LSTM are defined within the __init__() function and stored as local class properties. By storing key model attributes as local variables rapid alterations to the architecture can be made by re-initialising the object with a new set of starting parameters. Next a single forward pass and accompanying logic is defined within the forward() function [48], ordering the execution of the previously defined layers. The nn.LSTM layer [46] prediction internally computes the prediction from L stacked layers, which is next flattened and passed to a fully connected output layer nn.Linear [47]. The output is returned in addition to the hidden state to be used in the following recursive prediction.

The default Mini-Batch Gradient Descent methodology included within PyTorch was utilised, alongside PyTorch CUDA implementation, enabling fast parallelised training on a GPU. Training was performed on a Nvidia RTX 3070ti GPU consisting of 6144 CUDA cores and 8GB of VRAM. A DataLoader [49] was employed to iteratively load batches of 128 sequences to the GPU, which was limited by the size of the VRAM. Every model was trained for a duration of 20 epochs, with a limited test subsequently performed. This employed a randomised constrained testset (N=1000 sequences) as a heuristic of test loss. This methodology facilitated the identification of overfitting by comparison of the Training loss and Test loss, making it fundamental to the training of large models prevalent to overfitting. ADAM optimiser with a default learning rate of 0.001 was incorporated to dynamically reduce the learning rate. This enabled smaller changes to be made by the Mini-batch Gradient Descent Algorithm, fine tuning parameters to the identified minimum.

## 4.3.2 Hyper parameter Tuning

Evaluation of a broad spectrum of hyperparameter combinations was performed for this deep network. Hyperparameters such as number of layers and hidden memory cells are crucial aspects of the model and highly dependent on the complexity of the dataset. Accuracy and efficiency requirements are also a factor with increased hidden memory cells able to store more greater context, but significantly elevate computation requirements and escalate the risk of overfitting.

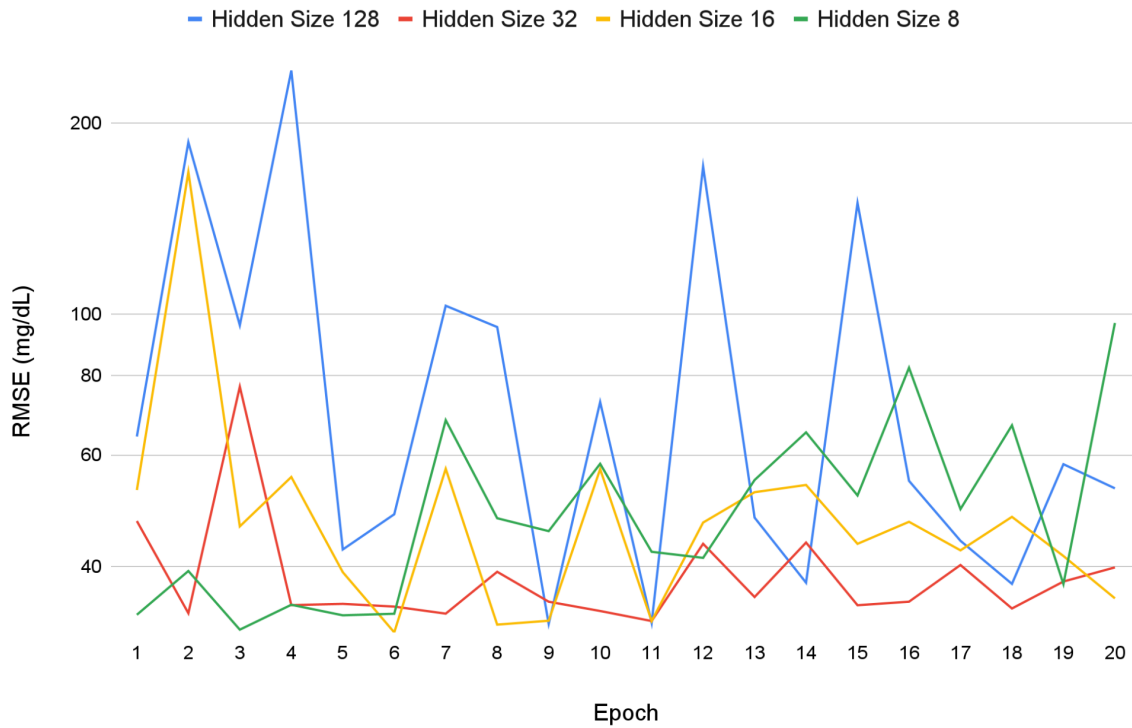

*Figure 15: LSTM RMSE loss on limited test set across variable hidden sizes*

Firstly, the effect of Hidden size on prediction performance was investigated.
Complexity of larger models was clearly illustrated by initial performance, where larger models had worse initial performance and required more epochs to reach minimum loss. Only 3 epochs were required for hidden size 8 to train, in contrast to 9 for 128 memory cells. Overfit was evident across larger models with rising discrepancies between train loss and test loss. This is clearly illustrated within epoch 4 for hidden size 128, which has the lowest training loss of 41.42 across all sizes, but the highest test loss of 242.00.
The best models: epoch 3 of hidden size 8 and epoch 6 of hidden size 16 were tested upon the whole testset, resulting in Root Mean Square Error (RMSE) of 33.31 and 33.07. Although the hidden size 16 has better performance, training took twice as long, which would be further exacerbated with the addition of stacked layers. Therefore 8 was chosen to be the optimal hidden memory size.

Secondly, the same approach was applied to investigate the layer depth.
A presumption can be made that the addition of stacked layers (Section 2.4.2) enabled deeper insights from a more abstracted viewpoint.

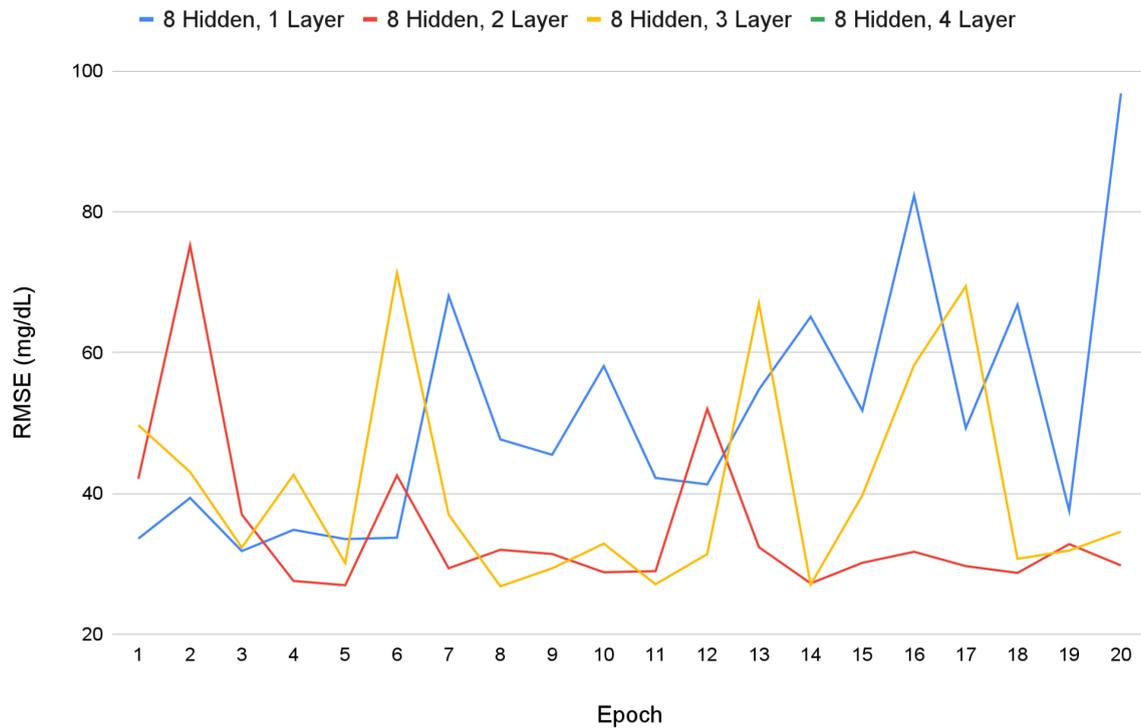

*Figure 16: LSTM RMSE loss on limited test set across variable amounts of layers*

Slight improvement over single layered vanilla LSTM, with the second layer making the most significant improvements to accuracy. Only marginal improvements were observed from the third layer, when compared to the addition of the second. Layer 4 did not coverage during training, having a consistent loss over 7300 and so it was inappropriate to test.

Epoch 5 of 2 stacked layers and epoch 8 of 3 layers were evaluated upon the whole test set, resulting in RMSE of 29.05 and 28.55 respectively. The optimal final model is made up of 1,513 trainable parameters and consists of 8 hidden memory cells, with 3 stacked layers. These hyperparameters were reused for all 5 repetitions of the cross-validation process.

# 5 Evaluation

## 5.1 Evaluation methodology

The evaluation will be conducted using a 5-fold cross-validation strategy (20% of the sequences being used to test and 80% to train), in addition to a random sampler, with a predefined seed to ensure repeatability. Sequences of consecutive measurements derived from a cohort, will be divided prior to segmentation and augmentation in order to eliminate duplicate sequences between the training and test sets. This eliminates the risk of test set contamination. Accuracy of the models will be evaluated with an array of metrics, for thorough comparative inspection of different loss types.

Root Mean Squared Error (RMSE) quantifies the proximity of predictions to the validation set. The PyTorch Mean Squared Error (MSE) loss function [50] used, had to be square rooted and then averaged over the number of batches in a training epoch.

Threshold Classification of glucose concentration measurements into hyperglycaemia, normal and hypoglycaemia, with thresholds of 70 mg/dL and 280 mg/dL, enabled the calculations of Precision and Recall metrics. Precision calculates the percentage of positive predictions given a real validation outcome. Similarly, Recall presents the relationship between real positive cases and predicted positive cases. Due to the possibility of serious consequences to patient health and life, an overestimation of false positives is most desireable. False negative predictions could delay appropriate rehabilitative action and exacerbate consequences. This creates a trade-off between precision and recall. A final calculation of the harmonic mean between precision and recall, known as the $F_1$ score [35], is used to quantify an unbiased measure impartial to precision and recall manipulation.

$$ESOD_n = \frac{ESOD(\hat{y})}{ESOD(y)}$$

$$= \frac{\sum_{k=3}^{N}(\hat{y}(k) - 2\hat{y}(k-1) + \hat{y}(k-2))^2}{\sum_{k=3}^{N}(y(k) - 2y(k-1) + y(k-2))^2}$$

*Equation 3: Formula definition of ESOD metric, given set of predictions ($\hat{y}$) and validation set (y)*

Energy of Second Order Difference (ESOD) reflects the risk of false alerts, indicating the reliability of a model [36]. The ideal ESOD score is 1, meaning the prediction is identical to the validation measurement.

## 5.2 Training Comparison

Evaluation was independently conducted on all four LSTM models, trained on both generalised and personalised methodologies, in order to recognise the best training procedure. This was conducted on only one of the 5-fold datasets.

|  | LSTM Model | RMSE Cohort 1 | RMSE Cohort 2 | RMSE Cohort 3 | Total RMSE |
|---|---|---|---|---|---|
| Cohort | Cohort 1, epoch 2 | 30.92 |  |  |  |
|  | Cohort 2, epoch 2 |  | 28.81 |  | **59.30** |
|  | Cohort 3, epoch 7 |  |  | 77.29 |  |
| All | All, epoch 8 | 30.51 | 27.62 | 27.87 | **28.52** |
| *Difference* |  | -0.41 | -1.19 | -49.42 | -30.78 |

*Figure 17: Comparison of RMSE between Training Sets*

A generalised training procedure is shown to yield higher performant models across the whole population when compared to a personalised methodology. Limited data availability used to train cohort-specific models impacted performance, with the largest cohort 1 (86 patients) having the smallest performance impact of -0.41 RMSE, whereas the smallest cohort 3 (31 patients) has a substantial accuracy loss of -49.42 RMSE. Performance could be further degraded for cohort 3 specific model, which has the lowest mean annual income, as this population was identified to have the greatest glucose concentration variability in section 3.2.3.

Vital to note possible test set contamination as the test sets created for each cohort were independently cross-validated from the generalised test set, meaning some sequences used within the generalised training set could be present in the cohort-specific test sets used to evaluate the generalised model.

Further investigations could be made into adapting the generalised model with transfer learning for possible decreases in loss for specific cohorts. However significant data set expansion would be necessary in order to avoid test set contamination, therefore inappropriate for the current available data.

## 5.3 Results

LSTM model was tested on the augmented test set, whereas the HMM and Heuristic models were tested on just a segmented test set. Testing of LSTM and HMM model was arduous as a result of the recursive prediction implementation , which introduced dependencies on previous prediction and memory state, meaning parallelisation was inhibited.

| Model | RMSE (mg/dL) | ESOD |
|---|---:|---:|
| Heuristic - Copy Last | 33.75 (6.85) | 0.00 |
| Heuristic - Linear Regression | 43.30 (7.31) | 1.81E-28 |
| HMM | 112.34 (15.73) | 7397.38 |
| LSTM | 28.55 (3.28) | 0.12 |

*Figure 18: Tabular evaluation of average model performance with corresponding s.d.*

LSTM network outperformed all other architectures at the 60-minute prediction interval, presenting excellent performance at predicting hypoglycaemia and hyperglycaemia, with precision of 0.865, recall 0.183 and $F_1$ score 0.303, in comparison to baseline Heuristic and HMM models. Given the large discrepancies between precision and recall, it is evident that thresholds require adjustments to be appropriate for medical use, with reasons outlined in Section 5.1. Further analysis focused on LSTM performance.

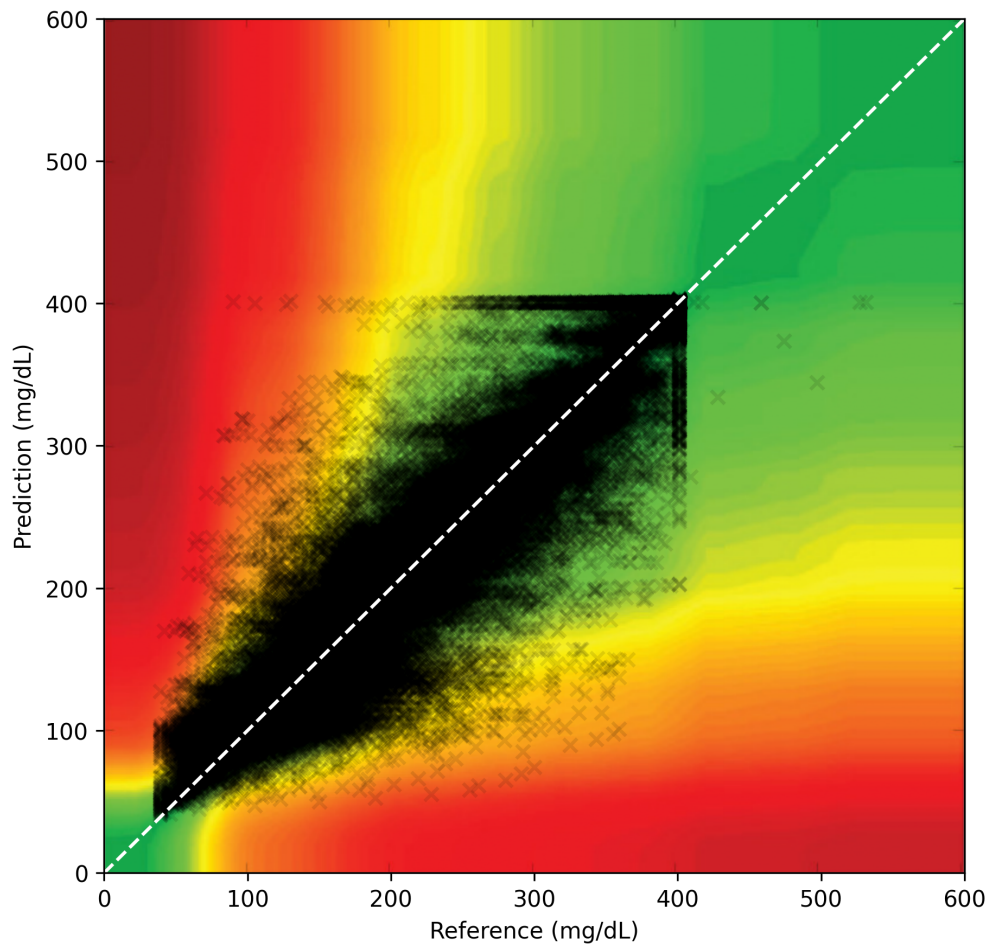

*Figure 19: Illustration of reference CGM readings (mg/dL) against LSTM predictions (mg/dL)*

The Surveillance error grid [78] quantifies accuracy by plotting predictions and reference values and visually comparing against the coloured continuous risk error gradient.
This is similar to Clarke Error Grid Analysis, however a continuous risk error grid is used as opposed to predefined zoning techniques [38]. It is evident that reference extrema readings over 400 mg/dL were never predicted by the LSTM model. This could be a result of severe under-representation in the dataset.

Predictions over yellow and orange areas, indicate elevated risk stratification, which poses an increased likelihood of health consequences. To add to that, subtle underestimation can be observed. This is highly agnostic in medical applications, as overestimation leading to unnecessary treatment is a more favourable, less dangerous scenario.

## 5.4 Model Comparison

|          | RMSE (mg/dL) | Dataset                    |
|----------|--------------|----------------------------|
| HMM      | 112.34       | CITY                       |
| LSTM     | 28.55        | CITY                       |
| LSTM [36]| 55.34        | OhioT1DM                   |
| LSTM [79]| 16.94        | The Maastricht Study (T2D) |
| ARIMA [79]| 27.77       | The Maastricht Study (T2D) |

*Figure 20: Comparison with State-of-Art Models (*subject to conversion)*

Other studies have also demonstrated successful recursive predictions of glucose levels using LSTM models. However direct comparison is not appropriate as a result of dataset discrepancies, with the OhioT1DM having significantly limited amounts of data and the Maastricht Study using data from Type 2 Diabetes patients.

## 5.5 Explainability

Deeper analysis with the aim of explaining decision processes, will be conducted upon the best performant model. LSTM model explainability can be investigated by assigning an importance to each datapoint in the time series. This could lead to the identification of trends in which events are significant, allowing medical professionals and patients to gain a deeper understanding of the reasoning behind each prediction. Furthermore there is potential for the identification of a patient's detrimental behaviour that increases risk of an hypo/hyperglycemic event occurring and should be altered. As featured in Section 2.5, mapping values of the forget gate vector indicates importance of the previous cell state [80].

Predictions based on training data have been outlined with the green line and those made recursively from previous predictions in red.

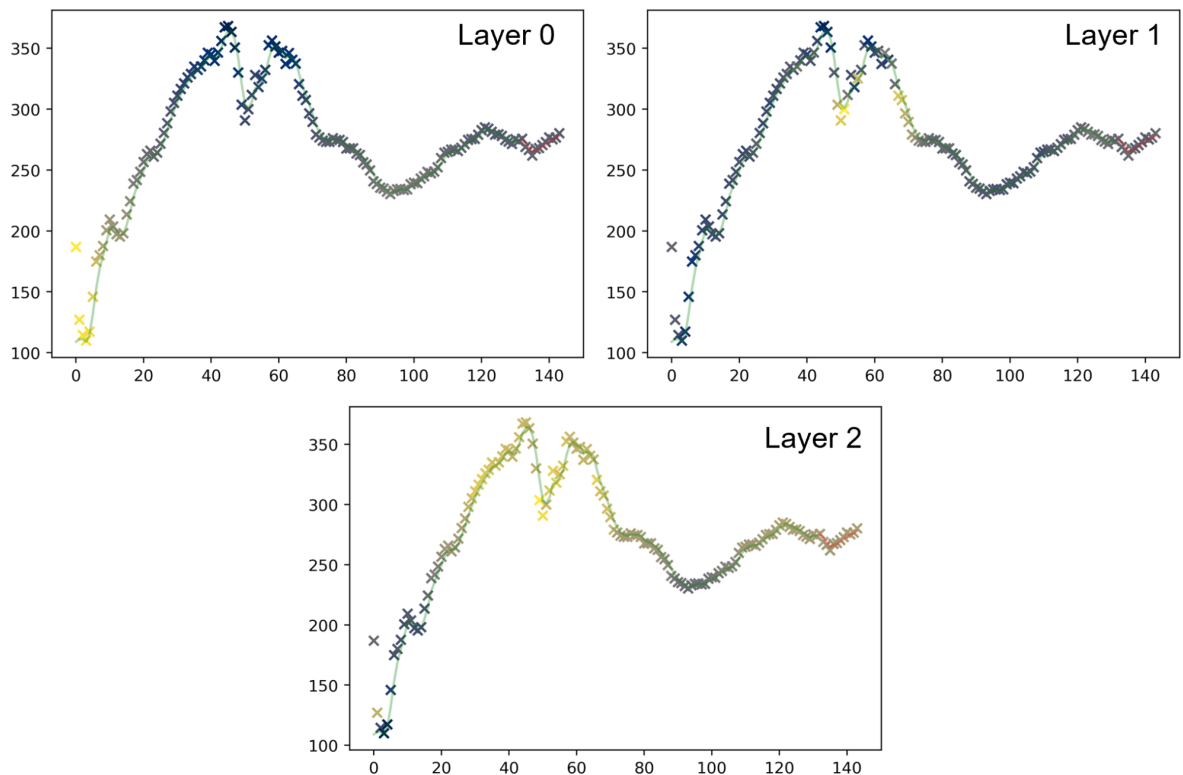

Figure 21: Importance of each CGM reading (mg/dL) in sequence, throughout all LSTM layers

It is clear elevated attention is placed on initial predictions as the hidden states are empty and require assignment.

The initial LSTM layer only utilises 2 out of 8 hidden states at indexes 0 and 3 and all others remain at a constant value of 1. A negative correlation is observed, with importance placed on the lowest glucose concentrations through the day. This could be interpreted as a basic identification of hypoglycemic events.

The first deep LSTM layer extracts more meaning, using all 8 hidden states.
Mean values increase around large variations of glucose levels, leading to the conclusion that focus is placed on the identification of large glucose concentration fluctuations throughout the day.

The next deep layer derives further meaning by using all hidden states.
Although the presented correlations are not as evident as previously, it can be deduced that importance is placed on high glucose values, which could improve predictions of hyperglycemia, unaccounted for by the first layer.
Furthermore as this is the final LSTM layer, previous to the fully connected layer, its values follow the desired output, and therefore positively correlate with CGM readings.

Investigating the explainability of the 2 stacked layer LSTM model, yielded similar findings and given the insights into similar loss between the 2 and 3 layered LSTMs (Section 4.3.2), the theory that the third layer had very minor contributions is reinforced.

It is vital to note the limitations of this method, which can only analyse a single sequence and requires visual inspection for each. This means it is not possible to quantify the trends observed in each layer, therefore comparison across all sequences in the test set is severely impractical for large datasets.

## 5.6 Limitations

The CITY dataset used demonstrates bias towards white and educated patients, as analysed in Section 3.2.1, so the cohort is unrepresentative of the American population.
This does not take into account the elevated demographic variations between countries such as the United States and the United Kingdom. Studies of other populations in India, have identified "high-calorie/high-fat and high sugar diets" [81], altering blood glucose concentrations when compared to others. This could lead to deficient accuracy in marginalised communities, which are not accurately represented in the patient cohort.

This could raise the need for personalised model trained at the state level to be better representative of the demographic population. However alternative approaches should be considered as patient clustering into smaller cohorts was shown to have a detrimental effect on performance.

# 6 Conclusions and Future work

## 6.1 Project Management

Given the research based nature of this project, covering a range of topics, significant planning was required. By restricting time for certain tasks, the scope of the research was limited and reducing the risk of overruns, therefore focus could be placed on the success criteria. Appendix B presents the planned work schedule, whereas Appendix C shows the schedule that took place.

Initial planning was formulated into a conventional waterfall methodology, with linear progression of tasks from data acquisition to training set creation. Simultaneous research was conducted in order to follow best practices, in addition to informing the ideation of the design methodology. This workflow was followed until 13th December 2021, where development was paused over the festive period, to allow more time for exam preparations and vocational activities.

Subsequent technical development from 31st January 2022 altered planning schemes to an Incremental Model, which appropriately divisioned end-to-end development of each model into separate linear progressions, from creation to evaluation. The perceived most challenging model: LSTM, was developed first, with Heuristics and HMM ensuing. Evaluation and Model comparison were performed at the conclusion of all model development. Concurrent report writing during model creation, enabled active recording of ideas and steps taken, minimising the risk of forgetting details.

Later focus was placed on formal documentation of the project, through the writing of this report. Visualisations were created using the python packages Matplotlib and Seaborn, in order to enhance and assist the conveyance of complex concepts. Additional Analysis was also necessary to assist with explanations.

## 6.2 Conclusions

Creation, Evaluation and Explainability Analysis was conducted on a range of techniques for predicting BG levels up to a 60 minute interval. Section 5.4 demonstrates LSTM model accuracy to be akin to state-of-art models in similar use cases. However as illustrated in figure 19, inaccurate predictions persist, elevating risk of hypo/hyperglycemia events. Therefore it is not considered appropriate for use in Artificial Pancreas applications. Nonetheless meaningful guidance is still provided to T1D patients.

Study has established a thorough training methodology for future Machine Learning models. Generalised models are proven to outperform personalised models (Section 5.2). Analysis of LSTM layers has shown the importance placed on major fluctuations in glucose concentration, as well as unconventional perspectives of high or low values.

## 6.3 Future Work

Independent peer review would be necessary for academic integrity and reproducibility of this approach which is crucial in the medical industry.

Future investigations could be conducted with addition of variable patient statistics such as weight, age and annual income directly to the model to provide further context.

An opportunity for deployment of the LSTM model in edge devices could arise, due to its small size and low energy requirements for inference. Sensitive medical data could be stored locally on the device and predictions computed locally [82]. Differential Privacy [84] techniques could also be implemented, leveraging the hidden state of LSTM to encode raw input data. This would allow data to be anonymously stored in the cloud without compromising the patient's privacy, as well as allowing for more detailed replication of prediction and deeper analysis of model error.

Furthermore, RMSE loss and other diagnostics may also be freely transmitted to web storage to aid evaluation, as it will not make patients uniquely identifiable. Similar approaches have already been implemented on Natural Language Processing of news articles using LSTM models for stock prediction [85]. Apprehensive patients could be persuaded by privacy enforcement, promoting patient cohort expansion and model invariance.

Compliance with regulatory policies [83] is vital for successful deployment to medical devices. These require extensive evaluation of each model, previous to certification, therefore continuous training and personalisation would not be possible.

# Appendices

## Appendix A: Project Brief

Artificial Intelligence has made large advances in preventive medicine and management of chronic conditions, such as Diabetes. The symptoms of this metabolic disease include elevated blood glucose levels and hyperglycaemic events, which over time can lead to irreversible damage to the heart, kidneys, blood vessels, eyes and nerves.

Type 1 Diabetes Patients must use external insulin, with effective management requiring optimal doses of insulin, through injection or infusion, multiple times per day.
Current technologies have found success in minimising risk of long-term complications, however they can pose a significant burden on self-managing patients, reducing their quality of life.

Recent growth in digital disease management platforms and an increased use of diabetes-related data sensors, such as Continuous Glucose Monitoring (CGM), have resulted in rapid evolution towards Autonomous Diabetes Systems, which implement Machine Learning approaches to predict future glucose levels, detect hyperglycaemic events, assist with decision making and risk stratification for Type 1 Diabetes patients.

The implementation of leading Neural Network based Machine Learning models could advance prediction accuracy and provide more personalised predictions. This leads to the opportunity to explain model predictions and justify conclusions reached, crucial for medical use.

Outlined below is my plan of approach:

1. Identify suitable Dataset for CGM use on patients with Type 1 Diabetes or combination thereof.
2. Extract and manipulate to generate adequate training and validation datasets.
3. Investigate using a variety of machine learning models
4. Evaluation and comparison of each model between each other as well as with the current leading predictive solutions for Autonomous Diabetes Systems.

Definition of success for the project:

- Creation of an accurate Neural Network Model to forecast glucose levels for patients with Type 1 Diabetes.
- Explain and justify the model's decision making process behind the predictions made.

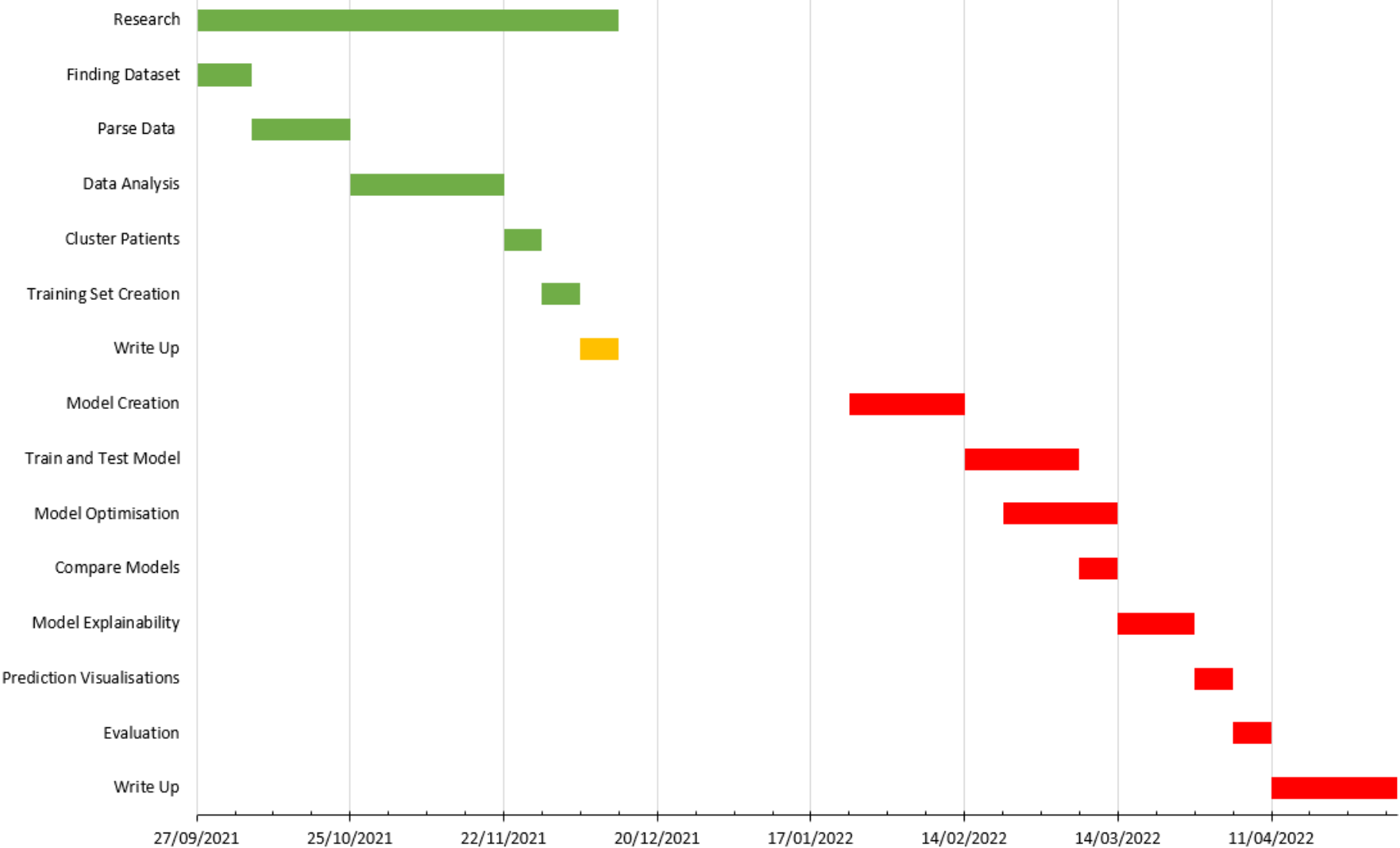

Appendix B: Planned Gantt Chart Brief

## Appendix C: Actual Event Gantt Chart

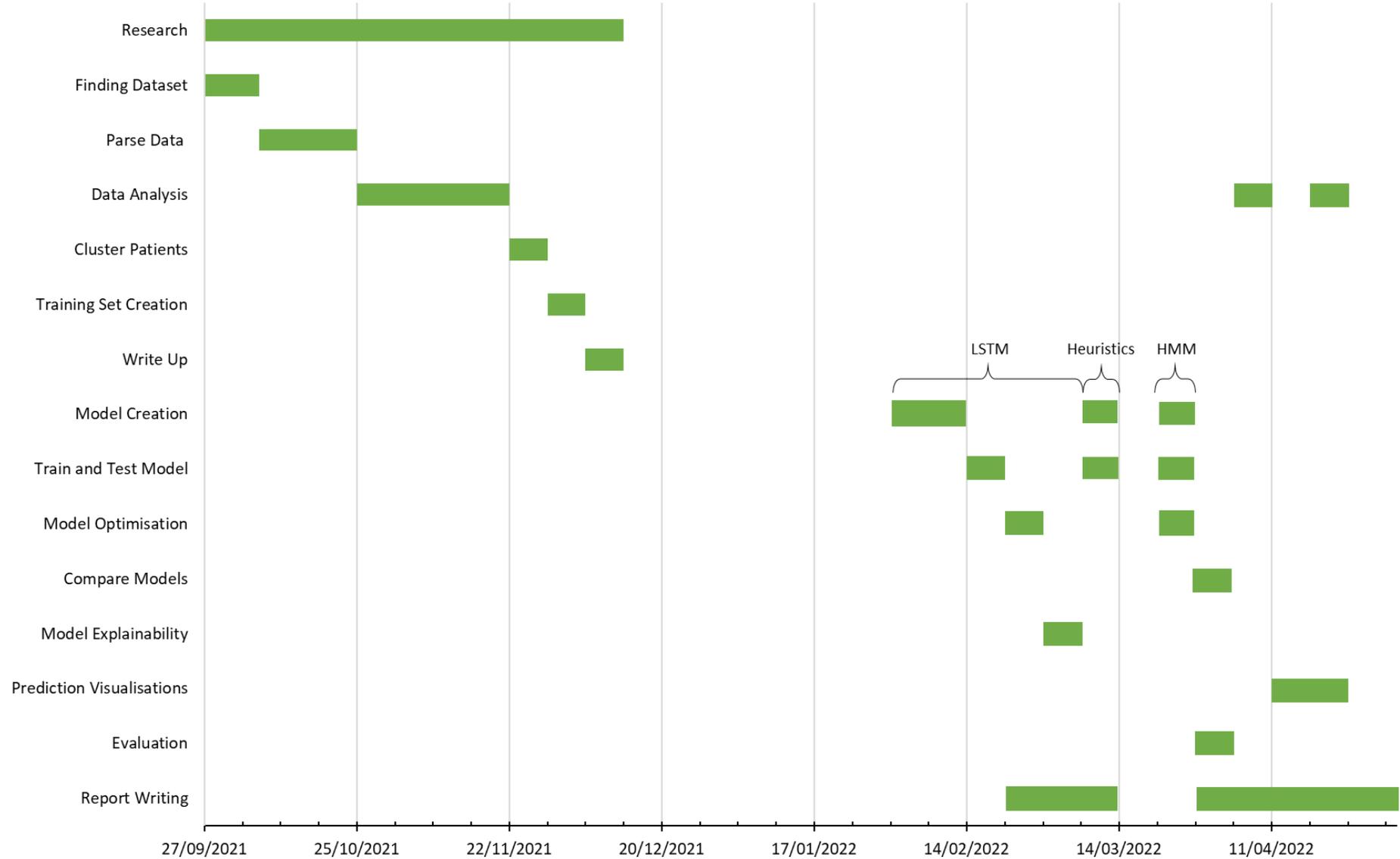